# Discovering Imperfectly Observable Adversarial Actions using Anomaly Detection

Olga Petrova
Avast Software
olga.petrova@avast.com

Karel Durkota
Dept. of Computer Science, FEE,
Czech Technical University in Prague
karel.durkota@fel.cvut.cz

Galina Alperovich
Avast Software
galina.alperovich@avast.com

Karel Horak
Dept. of Computer Science, FEE,
Czech Technical University in Prague
karel.horak@fel.cvut.cz

Michal Najman
Avast Software
michal.najman@avast.com

Branislav Bosansky
Dept. of Computer Science, FEE,
Czech Technical University in Prague
Avast Software
bosansky@fel.cvut.cz

Viliam Lisy
Dept. of Computer Science, FEE,
Czech Technical University in Prague
Avast Software
viliam.lisy@fel.cvut.cz

## ABSTRACT
Anomaly detection is a method for discovering unusual and suspicious behavior. In many real-world scenarios, the examined events can be directly linked to the actions of an adversary, such as attacks on computer networks or frauds in financial operations. While the defender wants to discover such malicious behavior, the attacker seeks to accomplish their goal (e.g., exfiltrating data) while avoiding the detection. To this end, anomaly detectors have been used in a game-theoretic framework that captures these goals of a two-player competition. We extend the existing models to more realistic settings by (1) allowing both players to have continuous action spaces and by assuming that (2) the defender cannot perfectly observe the action of the attacker. We propose two algorithms for solving such games – a direct extension of existing algorithms based on discretizing the feature space and linear programming and the second algorithm based on constrained learning. Experiments show that both algorithms are applicable for cases with low feature space dimensions but the learning-based method produces less exploitable strategies and it is scalable to higher dimensions. Moreover, we use real-world data to compare our approaches with existing classifiers in a data-exfiltration scenario via the DNS channel. The results show that our models are significantly less exploitable by an informed attacker.

## 1 INTRODUCTION
Anomaly detection is a general machine learning technique often used to find anomalous data, defects in products, or breakdowns of machinery. However, anomaly detection is also commonly used to detect malicious behavior, such as intrusions in computer networks [12], fraud in financial transactions [1], or malicious behavior of software [8]. In these domains, the effects of adversarial actions are being examined. The attackers execute their actions to optimize two aspects: (1) to not be detected by the anomaly detector and (2) to achieve their malicious goal (e.g., exfiltrate data from a computer network). From this perspective, using standard anomaly detection methods ignores the second aspect optimized by the attackers.

To explicitly reason about the goals of other agents, the game-theoretic framework is often used. There are several existing works where anomaly detection is integrated into a game-theoretic framework [6, 9, 10]. However, to the best of our knowledge, the existing game-theoretic models have two or more of the following limitations: They *(i)* assume discrete feature space over which the classification occurs; *(ii)* have very limited scalability with increasing dimensionality of this feature space; *(iii)* do not allow direct control of the false positive rate; *(iv)* and assume perfect observation of actions performed by the attacker. As a result, they are not readily applicable to many real-world problems, such as service misuse detection and data exfiltration. In these practical cases, features used for anomaly detection are often continuous (e.g., amount of data, entropy, etc.) and the actions of the attacker are interleaving with actions of regular benign users (e.g., when the attacker uses regular service for uploading data such as Dropbox or Google Drive that are also being used by the regular user). For this latter reason, it is not possible for the defender to exactly determine the action of the attacker nor the reward of the attacker. The defender only observes a signal that can be a combination of a benign and a malicious action (e.g., the total amount of uploaded data within a time window).

In this paper, we address all mentioned limitations and formulate the problem of adversarial anomaly detection where: *(i)* we allow features for anomaly detection to be continuous; *(ii)* there are hard constraints on false-positive rates for the anomaly detector; *(iii)* the actions of the attacker are not directly observable by the defender. We show that satisfying the first two extensions is possible by extending the existing results [9]. We derive a closed-form solution for the case with a known value of the game and combine this solution with a binary-search algorithm to optimize the value of the game with respect to the desired false-positive rate. For the case with unobservable actions of the attacker, we give two different algorithms.

As a baseline solution based on the existing works in this domain (e.g., [10]), we provide a linear programming (LP) formulation to compute the optimal solution of a discretized version of the problem. However, as we show in the experimental evaluation, the LP is not able to solve problems where the feature space for anomaly detection has many dimensions. To this end, we extend a recent learning-based algorithm *exploitability descent* [17] and we show that such an algorithm can be complemented with false-positive constraints and thus used in the domain of adversarial anomaly detection (or adversarial machine learning). We call this algorithm Exploitability Descent for Adversarial Anomaly Detection (EDA). This approach avoids the discretization step and approximates the optimal strategy of the defender by a neural network (NN). EDA ensures that the neural network is trained to make the attacker indifferent among his best options and that it meets the false-positive rate constraint.

The experimental evaluation on synthetic data shows that LP and NN produce very similar results in one or two-dimensional problems. However, any discretization-based approach cannot scale to larger dimensions. In order to compute some approximation within computation-time and memory limitations, we use increasingly sparse discretization points for increasing dimensions. In this comparison, the NN-based approach quickly outperforms the LP by producing more robust strategies for the defender. Second, we perform the experiment with real-world data of DNS queries that can be used for data exfiltration. We used a simple setting with only 3 features over the queries and show that our proposed method is significantly more robust compared to classical anomaly detectors that do not take the goals of the attackers into consideration. Compared to the detector computed by the proposed method, the attacker could gain 50% higher throughput of exfiltrated data against the best classical anomaly detector.

## 2 RELATED WORK

Widespread use of machine learning in security domains inspires analyzing their robustness against attackers. Empirical studies have shown that standard machine learning algorithms are vulnerable to well-tailored attacks [21, 24–26, 28]. It led a new field study – *adversarial machine learning* (AML). Surveys of the earlier works in AML for security domains are in [3, 14] and the more recent works are summarized in [27].

Research in AML consists of efficient *generation of adversarial examples* [11, 19] and designing machine learning *models robust against manipulation*. In this paper, we focus on the latter. One class of approaches for creating robust classifiers *obfuscates model gradients* in order to make searching for adversarial examples difficult. However, these approaches have been shown ineffective and easy to circumvent [2]. Another class, called *adversarial training*, modifies the training process to use the standard training data in combination with repetitively generated adversarial examples [13, 18, 25]. These approaches are similar to the method presented in this paper, but they do not have clear theoretic foundations and do not allow explicit consideration of players' incentives.

Recent years brought several principled approaches to AML. If the attacker tries to avoid detection without other preferences over the produced data samples, the problem may be formulated using *robust optimization* techniques [29]. In the case of a more complex structure of incentives, the framework of *game theory* is more suitable. Game theory provides a range of tools for computing optimal solutions in discrete game models. Hence, existing methods often relay on discretization [9, 10]. On the other hand, feature space discretization leads to computational complexity exponential in its dimensionality, which severely limits scalability. It can be partially resolved by training the models disregarding the attackers and making them robust by strategic selection of low-dimensional operating points [16]. This is sub-optimal since the models are not trained directly for the solved task.

Game-theoretic methods working directly with continuous feature spaces either introduce very limiting assumptions on the game structure [5, 6], or approximate behavior of discrete algorithms by neural networks and lose theoretic guarantees [22]. We follow the second approach. Based on the structure solutions introduced in [9], we propose an optimal game-theoretic solution for the simplified problem with false positive constraints, but perfect observation of the attacker's actions. We then extend this solution to the novel setting with uncertain observations using neural networks. To evaluate the performance of this extension, we propose an LP formulation based on [10] extended by kernel-based estimation of the benign data distribution.

## 3 PROBLEM SETTING

The interaction between the defender and the attacker can be modeled as a two-player game. The defender observes events that correspond to points in $n$-dimensional feature space and needs to decide whether to classify them as malicious. This decision can be randomized, and hence we consider that the classifier outputs a probability of a positive classification. The attacker observes the classifier and chooses an action which (1) generates an observation for the defender, and (2) provides the attacker with a reward in case the action goes undetected. For simplicity of the presentation, we assume that every action of the attacker can be associated with a point in the feature space (e.g., observation the defender would have received if he was able to separate effects of attacker's action from the effects caused by benign users), and that for every point in the feature space, there exists an action that maps to that point. This allows us to assume that the attacker directly chooses a point $f_a$ in the feature space (however, this point is not directly observed by the defender). To emphasize this distinction, we use $f_a$ for the actions of the attacker, and $f$ for the feature vectors observed by the defender.

Formally, we define a *Generalized Classification Game* as a tuple $G = (\mathcal{F}, C, R, P_D, \phi, \mathcal{T})$ where $\mathcal{F}$ corresponds to $n$-dimensional real-valued feature space corresponding to action space of the attacker and $C$ corresponds to a set of all classifiers that can be selected by the defender. Each classifier $c \in C$ corresponds to a classification function $c : \mathcal{F} \to [0, 1]$, and $c(f)$ is the probability that an event represented by a feature vector $f = (f^1, \ldots, f^n)$ gets inspected by the defender. We assume that each feature $f^i$ is bounded, and we denote these bounds by $[L^i, U^i]$. In case the attacker executes action $f_a \in \mathcal{F}$ and does not get detected, he receives a non-negative reward $R(f_a) \geq 0$. The inspection capacity of the defender is limited, and a maximum false-positive rate $\phi \in [0, 1]$ of classifying benign events as malicious is allowed. We assume that the benign events are samples from a distribution $P_D$, hence the expected false-positive rate of a classifier $c$, denoted $\Phi_D(c)$, satisfies $\Phi_D(c) = \mathbb{E}_{f \sim P_D} c(f)$.

Finally, $\mathcal{T}$ corresponds to an observation transformation function. For a given action of the attacker, associated with a feature vector $f_a \in \mathcal{F}$, the defender may observe feature vectors sampled from a probability distribution $\mathcal{T}(f_a) \in \Delta(\mathcal{F})$.[1] In general, we do not pose any restrictions on the transformation function $\mathcal{T}$, but we focus on two specific cases in the study. First, we describe a simple case that corresponds to existing models where $\mathcal{T}$ reveals the action of the attacker to the defender, i.e., $\Pr_{\mathcal{T}}[f = f_a | f_a] = 1$. Second, we describe a case where the action of the attacker is additively combined with an action $f_b \sim P_D$ of a legitimate user, $\mathcal{T}(f_a) = f_a + P_D$, i.e., $\Pr_{\mathcal{T}}[f = f_a + f_b | f_a] = \Pr_{P_D}[f_b]$.

*Example 3.1.* Consider a case where the attacker wants to upload a 100 MB file from an infiltrated computer. To hide the activity, the attacker is using a Dropbox folder for the upload. The goal of the attacker is to upload the file as quickly as possible, however, he observes that the standard user of the infiltrated computer uses Dropbox very rarely and the user never uploads large files. The attacker knows that an anomaly detection system is running on a network and data over every 1 hour are aggregated and evaluated for anomalies. If during this 1 hour time window both the user and the attacker upload data to Dropbox, the aggregated statistic shows the sum of these two uploads (function $\mathcal{T}$). Therefore, the attacker has to consider the behavior of the user (in a form of approximating a probability distribution $P_D$) and choose the amount of the uploaded data ($f_a$) such that if the user uploads data as well, the overall amount of data is unlikely to be detected as an anomaly.

*Expected Utility.* The attacker only gets a reward for executing action $f_a$ if he does not get detected by the defender. His expected utility when executing action $f_a$ against classifier $c$, denoted $u_a(c, f_a)$, hence satisfies

$$u_a(c, f_a) = (1 - \rho_c(f_a)) \cdot R(f_a), \qquad (1)$$

where $\rho_c(f_a) = \mathbb{E}_{f \sim \mathcal{T}(f_a)}[c(f)]$ is the expected probability that $c$ classifies action $f_a$ (based on observations $f \sim \mathcal{T}(f_a)$) as anomalous. We assume a zero-sum game setting where the goal of the defender is to minimize the expected reward of the attacker.

The solution of the Generalized Classification Game corresponds to a maximin (or a Stackelberg equilibrium) where the defender chooses a classifier $c$ first and the attacker follows by playing a best response against the classifier, i.e., action $f_a$ that maximizes attacker's expected utility $u_a(c, f_a)$. Formally, the defender seeks a classifier $c^*$ that minimizes the attacker's utility under a false-positive rate constraint *assuming* that the attacker plays a best response:

$$c^* = \arg\min_{c \in C} \ \{\max_{f_a \in \mathcal{F}} u_a(c, f_a)\} \qquad (2a)$$

$$\text{s.t.} \ \Phi_{P_D}(c) \le \phi \qquad (2b)$$

We use $BR_a(c)$ to denote the best response of the attacker to the classifier $c$, where $BR_a(c) = \arg\max_{f_a} u_a(c, f_a)$.

---
[1] Recall that the feature vector $f_a$ corresponding to attacker's action cannot be observed and used for the classification.

## 4 IDENTITY OBSERVATION TRANSFORMATION FUNCTION

Before discussing the more general setting, we first describe the baseline case where $\mathcal{T}$ is an identity and that corresponds to previous work [9] that operated over a discrete feature space:

$$\Pr_{\mathcal{T}}[f | f_a] = \begin{cases} 1 & f = f_a \\ 0 & f \ne f_a. \end{cases} \qquad (3)$$

In this case, the defender can directly observe the attacker's action $f_a$ and thus also the attacker's reward $R(f_a)$. In order to obtain a robust anomaly detector, we optimize the classification strategy $c : \mathcal{F} \to [0, 1]$ against the worst-case attacker. Such an attacker aims to choose the action with the highest expected reward for the attacker $u_a(c, f_a) = (1 - c(f_a))R(f_a)$. Since the action of the attacker is observable, we define the classifier directly on the attacker's actions, and we write $c(f_a)$ instead of $c(f)$.

Following Dritsoula et al. [9], we formulate a simple algorithm that solves this game. First, assume that the value $V^* \ge 0$ of the game is known. Then, the utility of *any* action $f_a$ of the attacker must be less than or equal to $V^*$,

$$(1 - c(f_a))R(f_a) \le V^* . \qquad (4)$$

This yields a lower bound on the probability an action $f_a$ of the attacker has to be inspected by an optimal classifier $c^*$ to achieve (at least) the value $V^*$,

$$\rho_{c^*}(f_a) \ge \max\{1 - V^*/R(f_a), 0\} . \qquad (5)$$

Since the defender is able to observe attacker's action $f_a$, the probabilities $\rho_{c^*}(f_a)$ and $c^*(f_a)$ coincide. Hence, an optimal classifier $c^*$ can be constructed as $c^*(f_a) = \max\{1 - V^*/R(f_a), 0\}$.

We have shown that in the case $V^*$ is known, we can represent the optimal classifier using a closed-form solution. In practice, the value $V^*$ of the game is, however, unknown. To find it we propose an approach termed Closed-Form based Search (CFS) algorithm shown in Algorithm 1. Denote CFS($V$) a classifier that achieves value $V$ using Eq. (5), i.e.,

$$\text{CFS}(V)(f) = \max\{1 - V^*/R, 0\} . \qquad (6)$$

The Algorithm 1 uses bisection method to locate $V^*$ within an interval $[\underline{V}, \overline{V}]$. Initially, we set $\underline{V} = 0$ and $\overline{V} = \max_{f \in \mathcal{F}} R(f)$. Value $\underline{V} = 0$ is achieved by a classifier CFS($\underline{V}$) that classifies each $f_a$ with $R(f_a) > 0$ with probability CFS($\underline{V}$)($f$) = 1 and thus possibly has excessive false-positive rate $\Phi_D(\text{CFS}(\underline{V})) > \phi$. In contrary, value $\overline{V}$ is achieved by a classifier CFS($\overline{V}$) which classifies all events as benign. To minimize the attacker's utility, we search for $V^*$ such that false-positive rate of classifier CFS($V^*$) is maximized and satisfies $\Phi_D(\text{CFS}(V^*)) \le \phi$.

In each iteration, the algorithm considers the middle point $\hat{V}$ of the currently considered interval $[\underline{V}, \overline{V}]$ and inspects the false positive rate of the classifier CFS($\hat{V}$). If the false-positive rate $\Phi_D(\text{CFS}(\hat{V}))$ of the classifier higher than the desired rate $\phi$, the value $\hat{V}$ cannot be achieved within this false positive constraint. Hence the value of the game has to be higher (i.e., better for the attacker) and the algorithm searches interval $[\hat{V}, \overline{V}]$ on line 3. On the contrary, when the false-positive constraint is less than $\phi$, the algorithm searches the other interval $[\underline{V}, \hat{V}]$ trying to minimize the value of the game. When

**Algorithm 1:** Closed-Form based Search (CFS)

1  **while** $\overline{V} - \underline{V} \geq \epsilon$ **do**
2  $\quad \hat{V} \leftarrow (\underline{V} + \overline{V})/2$
3  $\quad$ **if** $\Phi_D(\text{CFS}(V^*)) > \phi$ **then** $\underline{V} \leftarrow \hat{V}$
4  $\quad$ **else** $\overline{V} \leftarrow \hat{V}$
5  **return** $\text{CFS}(\overline{V}))$

the desired $\epsilon$-approximation of $V^*$ is found, the algorithm returns classifier $\text{CFS}(\overline{V})$.

## 5 GENERAL OBSERVATION TRANSFORMATION FUNCTION

The key assumption of the previous case is that the defender perfectly observes the action of the attacker $f_a$, the defender thus exactly knows the reward of the attacker $R(f_a)$, and can apply Eq. (5) to calculate the probability with which the event has to be inspected. We now describe the case with a general observation transformation function $\mathcal{T}$ where the action of the attacker (and thus neither the reward) is observable by the defender and the approach described in the previous section cannot be applied.

Following our example, we primarily focus on the observation transformation function $\mathcal{T}$ that corresponds to a sum of effects of actions of the attacker ($f_a$) and the user ($f_b$). In this case, the actions of the attacker $f_a$ are additively combined with the actions $f_b \sim P_D$ of the legitimate users, i.e.,

$$\Pr_{\mathcal{T}}[f = f_a + f_b \mid f_a] = \Pr_{P_D}[f_b]. \quad (7)$$

Note that the behavior of the legitimate user is stochastic and unpredictable. The defender observes only a single data point $\mathcal{T}(f_a) = f_a + f_b$ and cannot easily infer the action of the attacker $f_a$ since the action of the user $f_b$ is also not known and only the distribution $P_D$ is known.

In this section, we propose two algorithms to solve classification games with general observation transformation functions $\mathcal{T}$ – one based on the discretization of action and feature spaces and linear programming, and the second one that approximates the optimal strategy using a neural network.

### 5.1 Solving the Generalized Classification Game - Linear Programming Approach (LP)

We first present the algorithm that is a natural extension of previous algorithms that solve the discrete cases [10]. The algorithm assumes a discretized feature space such that each dimension (feature) is divided into $d$ uniform steps (subintervals). We denote $\mathcal{F}_d$ the set of all the subintervals of the feature space (or grid cells). We use bar notation $\bar{f} \subset \mathcal{F}$ to denote one element of the discretized set such that $f \in \bar{f} \in \mathcal{F}_d$. Discretization of the feature space modifies the game such that both players have a finite number of actions – the action of the attacker corresponds to choosing a grid cell, and the defender chooses a probability of inspection of each grid cell. Let $\Pr[\bar{f}|A]$ denote the probability density in a cell $\bar{f}$ given the probability distribution $A$. Now, we can formalize a linear program based on standard LP for solving one-shot (matrix) games with finite sets of actions (e.g., from [23]):

$$\min_{\forall \bar{f}_o \in \mathcal{F}_d: c(\bar{f}_o)} V \quad (8)$$

$$\text{s.t.:} (\forall \bar{f}_a \in \mathcal{F}_d) : \hat{R}(\bar{f}_a) \sum_{\bar{f}_o \in \mathcal{F}_d} \Pr[\bar{f}_o | \hat{\mathcal{T}}(\bar{f}_a)](1 - c(\bar{f}_o)) \leq V \quad (9)$$

$$\sum_{\bar{f}_o \in \mathcal{F}_d} \Pr[\bar{f}_o | P_D] c(\bar{f}_o) \leq \phi \quad (10)$$

$$(\forall \bar{f}_o \in \mathcal{F}_d) : 0 \leq c(\bar{f}_o) \leq 1, \quad (11)$$

There are two types of variables – $V$, representing the value of the discretized game that as expected utility of the attacker, and $c(\bar{f}_o)$ for each $\bar{f}_o \in \mathcal{F}_d$ representing a probability with which an even from the grid cell $\bar{f}_o$ will be inspected by the defender. Constraints (9) correspond to best responses of the attacker, thus an expected value for each action of the attacker $\bar{f}_a$ is less or equal $V$. Since the feature space is discretized, we assume the attacker is choosing such an action from a grid cell $\bar{f}_a$ that maximizes his reward $f_a^m = \arg\max_{f_a' \in \bar{f}_a} R(f_a')$, thus $\hat{R}(\bar{f}_a) = R(f_a^m)$ and $\hat{\mathcal{T}}(\bar{f}_a) = \mathcal{T}(f_a^m)$. The strategy of the defender is a probability distribution over each grid cell to inspect, hence the values are bounded (11) and constraint (10) restricts the defense strategy to allow the maximal false-positive rate of $\phi$. While LP can be solved in polynomial time w.t.r. to the size of the input, in our case the input matrix size grows exponentially with dimensions $n$.

There are two main disadvantages of the LP-based approach. First, the scalability of the method is limited due to the size of the program. While this can be to some extent tackled by the incremental generation of constraints and/or variables, the exponential size of the LP is unavoidable in general. Second, the LP in the present form overfits the benign data. If the input for the program is a sample of data (in case we do not know the exact distribution), the strategy computed using the linear program and data samples $D$ can violate the false-positive constraints on a different data sample $D'$ from the same distribution.

To overcome this issue, we approximate the distribution $P_D$ from a data sample $D$ using Kernel Density Estimation (KDE) with kernel bandwidth parameter $h$. We find the best value for the hyperparameter $h$ using a binary search. Consequently, the approximated distribution is used for a false-positive constraint (10).

### 5.2 Exploitability Descent for Adversarial Anomaly Detection (EDA)

We now turn to the main algorithm inspired by *exploitability descent* [17]. We solve the min-max problem in Eq. (2a) iteratively by conducting two phases at each iteration: first, we estimate the best response of the attacker $f_a$ to the current classifier $c$ and, second, we update the classifier $c$ with stochastic gradient descent while taking into account the false-positive rate constraint. We model the classifier with a neural net $c_\theta$ which is parametrized by the weights $\theta$.

The adversarial nature of the problem makes the classifier tend to output inspection probabilities of the attack close to 1 and the inspection probabilities of the benign samples close to 0. Such values may cause the vanishing gradients problem due to the sigmoid in the last layer of the neural net [15, 20]. To partially overcome this issue,

**Algorithm 2:** Exploitability Descent for Adversarial Anomaly Detection (EDA).

**Input:** $\hat{V}, \phi$
1 $\theta^0 \leftarrow$ initial random NN
2 $\lambda^0 \leftarrow 1.0, i \leftarrow 0$
3 **while** *termination condition is not met* **do**
4 $\quad f_a \leftarrow BR_a(c^i)$
5 $\quad \hat{\phi}^i \leftarrow \Phi_D(c^i)$
6 $\quad \mathcal{L}^i \leftarrow -\mathbb{E}_{f \sim \mathcal{T}(f_a)}[\log(c^i(f))] + \lambda^i(\hat{\phi}^i - \phi)$
7 $\quad \theta^{i+1} \leftarrow \theta^i - \alpha \nabla_\theta \mathcal{L}^i$
8 $\quad \lambda^{i+1} \leftarrow \max\{0, \lambda^i + \beta \nabla_\lambda \mathcal{L}^i\}$
9 $\quad i \leftarrow i + 1$
10 $\theta \leftarrow OutputClassifier(\{\theta^0, \theta^1, \dots\}, \phi)$
**Output:** $\theta$

we conveniently construct an upper bound of the outer minimization problem of (2a) by taking the logarithm of the criterion and minimize the upper bound instead.

*Objective.* Given a sample of data $D$ and a best response $f_a \in BR_a(c_\theta)$, the optimal classifier $c_{\theta^*}$ is obtained by solving the following task:

$$\min_{\theta \in \Theta} \quad (1 - \mathbb{E}_{f \sim \mathcal{T}(f_a)}[c_\theta(f)]) \cdot R(f_a), \quad (12)$$

$$\text{subject to: } \Phi_D(c_\theta) \leq \phi \quad (13)$$

Once we compute the best response $f_a$ in the first phase of each iteration of EDA, we consider it is fixed. Therefore, in the second phase, the reward $R(f_a)$ is not a function of the weights $\theta$ and it can be omitted from the criterion. This gives a simplified objective $-\mathbb{E}_{f \sim \mathcal{T}(f_a)}[c(f)]$. Note that the simplified objective is minimized by the same optimal classifier classifier $c_{\theta^*}$ as the following criterion $-\log \mathbb{E}_{f \sim \mathcal{T}(f_a)}[c_\theta(f)])$ due to monotonicity of logarithm. Per Jensen's inequality, we construct the upper bound on the simplified objective.

$$-\log \mathbb{E}_{f \sim \mathcal{T}(f_a)}[c_\theta(f)]) \leq -\mathbb{E}_{f \sim \mathcal{T}(f_a)}[\log(c_\theta(f))] \quad (14)$$

We further use Lagrangian relaxation procedure (described in Chapter 3 [4, 7]) to move the false-positive rate constraint into the objective to obtain unconstrained problem as follows:

$$\max_{\lambda \geq 0} \min_{\theta \in \Theta} -\mathbb{E}_{f \sim \mathcal{T}(f_a)}[\log(c_\theta(f))] + \lambda \cdot (\Phi_D(c_\theta) - \phi) \quad (15)$$

*Pseudocode.* The pseudocode of EDA is presented in Algorithm 2. The algorithm takes false-positive rate threshold $\phi$ and a set of benign data $D$ as an input. The neural network $c_\theta$ is parameterized by $\theta$. The output of the algorithm is an optimal classifier $c_{\theta^*}$ The parameters $\theta$ and $\lambda$ are updated in each iteration.

At each iteration, the algorithm estimates the best response of the attacker (line 4) against the current classifier $c_{\theta^i}$ and the current false-positive rate $\hat{\phi}^i$ given the benign data $D$ (line 4). Then, as in Eq. (15), the Lagrangian $\mathcal{L}^i$ is constructed by combining the upper bound and the constraint (line 6). Finally, we update the weights $\theta^i$ with a gradient descent step (line 7) and the multiplier $\lambda_i$ with a gradient ascent step (line 8).

## 5.3 Selecting the Final Classifier

At each iteration of the algorithm, a new classifier $c^i$ is constructed. Since the false-positive constraint is not strictly enforced during the learning, the false positive rate $\hat{\phi}^i$ of some of the classifiers $c^i$ may exceed the desired false-positive rate $\phi$, and thus become unacceptable as a solution of the problem. Although such classifiers exceed the threshold on false-positives, they may achieve a good value against the best response of the attacker $u_a(c^i, BR_a(c^i))$—and disregarding them completely can be wasteful. We show, however, that multiple classifiers can be linearly combined to obtain a new classifier that matches the desired false-positive rate, and that provides a superior classification performance.

PROPOSITION 5.1. *Let $c^i$ and $c^j$ be two classifiers such that the best-responding attacker achieves utilities $u_a(c^i, BR_a(c^i))$ and $u_a(c^j, BR_a(c^j))$, respectively. Let $\lambda \in [0, 1]$ and consider a classifier $c_\lambda^{ij}$ such that $c_\lambda^{ij}(f) = \lambda c^i(f) + (1 - \lambda)c^j(f)$, i.e., $c_\lambda^{ij}$ is a convex combination of classifiers $c^i$ and $c^j$. Then $c_\lambda^{ij}$ satisfies*

*(1) $\Phi_D(c_\lambda^{ij}) = \lambda \Phi_D(c^i) + (1 - \lambda)\Phi_D(c^j)$, and*
*(2) $u_a(c_\lambda^{ij}, BR_a(c_\lambda^{ij})) \leq \lambda u_a(c^i, BR_a(c^i)) + (1 - \lambda)u_a(c^j, BR_a(c^j))$.*

PROOF. The classifier $c_\lambda^{ij}$ corresponds to rolling dice before classifying and using the classifier $c^i$ with probability $\lambda$ (and $c^j$ otherwise). Hence the false-positive rate of $c_\lambda^{ij}$ satisfies

$$\Phi_D(c_\lambda^{ij}) = \mathbb{E}_{f \sim P_D}[c(f) \mid c = c_\lambda^{ij}]$$
$$= \mathbb{E}_{f \sim P_D}[c(f) \mid c = c^i]P[c = c^i]$$
$$\quad + \mathbb{E}_{f \sim P_D}[c(f) \mid c = c^j]P[c = c^j]$$
$$= \lambda \mathbb{E}_{f \sim P_D}[c(f) \mid c = c^i] + (1 - \lambda)\mathbb{E}_{f \sim P_D}[c(f) \mid c = c^j]$$
$$= \lambda \Phi_D(c^i) + (1 - \lambda)\Phi_D(c^j).$$

Similarly, if the attacker knew which one of the classifiers is used (i.e., the result of the dice roll), he would have been able to play the best responses $BR_a(c^i)$ and $BR_a(c^j)$, respectively, to achieve utility $\lambda u_a(c^i, BR_a(c^i)) + (1 - \lambda)u_a(c^j, BR_a(c^j))$. Since he does not have access to this information, his utility can only worsen, which proves (2). □

Based on Proposition 5.1, we compute the resulting classifier by finding the optimal pair of classifiers $c^i$ and $c^j$, and the optimal coefficient $\lambda$ of convex combination, such that $\lambda \Phi_D(c^i) + (1 - \lambda)\Phi_D(c^j) \leq \phi$ and the upper bound $\lambda u_a(c^i, BR_a(c^i)) + (1 - \lambda)u_a(c^j, BR_a(c^j))$ on the attacker's utility is minimized.

We represent the performance of the classifiers in the form of a 2-dimensional Pareto frontier, where each classifier $c^i$ is assigned a point $(u_a(c^i, BR_a(c^i)), \Phi_D(c^i))$. We also use this Pareto frontier to establish the termination condition of the Algorithm 2. Specifically, we terminate the algorithm when the Pareto frontier does not change for $K$ iterations, where we use $K = 50$ in the experiments.

*Learning Rates.* Finally, we discuss the learning-rate steps $\alpha$ and $\beta$ for updating weights of the neural network and lambda, respectively. We chose the values according to condition (12) from [7] such that the step decreases faster for the neural network and for $\lambda$. Since we use the same learning rates for neural networks of various sizes, we use a simple restarting mechanism whenever the false-positive value remains constant and within the distance of 0.01 from either value 0 or 1 for $L = 500$ iterations. In this case, the whole algorithm is restarted and the constants of learning rates are decreased by 10% while still satisfying the condition (12).

## 6 EXPERIMENTS

We now turn to the experimental evaluation of the algorithms. We first focus on synthetically generated data to examine the quality of computed strategies with varying parameters of the problem. We follow by a real-world scenario where data correspond to DNS requests and the goal is to identify anomalies caused by an attacker exfiltrating data using DNS. All algorithms were run on Intel Xeon Scalable Gold 6150 (2.7GHz) with 20 GB memory limit and 24 hours time-limit per instance. We use PyTorch to train the neural networks and Gurobi solver to solve LP problems.

In the experiments, we always consider three data sets: training $D_{train}$ (50% of all data points), validation $D_{val}$ (25% data points) and testing $D_{test}$ (25% data points). The training set is used to train the model, the validation set to find the model hyperparameters and thresholds (e.g., KDE bandwidth parameter $h$, thresholds for classical anomaly detectors, etc.). Finally, the testing data set is used for the final performance evaluation (and constitutes the reported results). We consider false-positive rate threshold $\phi = 0.05$.

*Parameters for EDA.* Based on the experimental evaluation, we use the following architecture of neural networks for the EDA algorithm. The neural networks consist of 3 fully-connected layers with $32 + 2n$, $32 + 2n$, $16 + 2n$, and 1 (output) neurons, where $n$ is the dimension of the feature space. We use the ReLU activation function for all neurons except the last one, where the sigmoid function is used.

We estimate the best response $BR_a(c)$ of the attackers (line 4 of the Algorithm 2) by sampling a set $S$ of $256n$ attacker's actions from $\mathcal{F}$. Each sampled action $f_a \in S$ is then improved using hill-climbing method and the sample $f_a \in S$ with the highest reward of the attacker $u_a(c, f_a)$ is used as an estimate of $BR_a(c)$.

*Method for Comparing Different Classifiers.* We are interested in the quality of classifiers that is determined by the exploitability – what is a value of the best response of the attacker against the computed model. Due to the size of the problem, the best response cannot be solved optimally and we can only estimate the value of a best response. To do so, we sample the feature space with a given number of samples ($2,000 \cdot n$ samples for the evaluation on synthetic data, 30,000 samples for the real-world data) and select the best value for the attacker (subsequent stochastic gradient descent is not used for this evaluation due to the discretization used for LP-based methods). Such a heuristic estimation of a best response does not have any theoretical guarantees. Therefore we experimentally evaluate the variance of the sampled best response. In Figure 1 we present a histogram of values of 10,000 estimations based on sampled best response with 30,000 samples used for experiments in Section 6.2. The results show that the variance of the sampled best response is rather small as the lowest value within these 10,000 estimations was 58.37 while the highest estimation was 58.75.

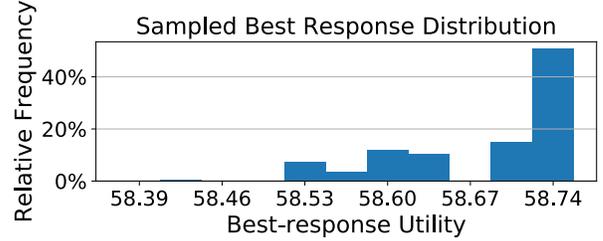

Figure 1: Evaluation of variance of sampled best response with 30,000 samples used for data exfiltration experiments.

| alg \ $n$ | 1 | 2 | 3 | 4 | 5 | 10 |
|---|---|---|---|---|---|---|
| LP $d$ | 1,000 | 80 | 12 | 5 | 4 | - |
| LP [s] | 0.1 | 5.6 | 247.8 | 424.3 | 1,885.7 | N/A |
| EDA [s] | 3,241 | 3,719 | 4,782 | 5,967 | 5,935 | 12,571 |

Table 1: Algorithms run time (in seconds) for different number of dimensions $n$. For LP, we used discretization $d$ according to the table.

### 6.1 Evaluation on Synthetic Benign Data

We use synthetic benign data to analyze our game-theoretic models and determine and compare the performance of described algorithms for varying parameters, focusing primarily on the dimension of the feature space. We create the datasets of benign data by sampling each coordinate $f_b^i$ of a feature vector $f_b = (f_b^1, \ldots, f_b^n)$ from normal distribution $\mathcal{N}(\mu^i, \sigma^i)$. Means $\mu_i$ are sampled from uniform distribution $U(1.6, 8.4)$ and the standard deviation $\sigma_i$ is sampled from a uniform distribution $U(1, 1.25)$. By repeating the process (for fixed $\mu^i$ and $\sigma^i$), we generate a dataset of $1000n$ samples. We consider that the reward function $R$ of the attacker is a sum of individual features from the feature vector $f_a$, $R(f_a) = \sum_{i=0}^{n} f_a^i$.

*6.1.1 Algorithm Scalability.* We first examine the scalability of the methods with the increasing dimension of the feature space. Table 1 presents the mean running times (5 runs). The number of discrete steps in each dimension is decreasing for the LP-based method in order to compensate for the exponential number of variables/constraints. The results show that while LP-based method is faster for smaller dimensions, the computation time required to construct the LP is increasing with the dimensions due to the computation of the density w.r.t. to KDE that is necessary for keeping the false-positive rate close to the desired value. Note that we were unable to run the LP-based method for $n = 10$ within 24 hours deadline. On the other hand, EDA was able to converge in all instances within the time limit, including the case with dimension $n = 20$ that took slightly more than 14 hours to converge on average. Note however that the majority of the computation time of EDA was spent in the best-response computation. Already for 3 dimensions, computation of best responses takes more than 93% of time and the fraction is further increasing for higher dimensions. By implementing a problem-specific best-response algorithm or using parallel (possibly GPU-based) best-response computation, the computation time of EDA can be significantly reduced.

|   | Identity $\mathcal{T}$ | | General $\mathcal{T}$ | |
|---|---|---|---|---|
| $n$ | LP[%] | EDA[%] | LP[%] | EDA[%] |
| 1 | 4.98±0.46 | 4.96±0.45 | 5.27±1.12 | 5.19±0.68 |
| 2 | 4.95±0.19 | 5.02±0.30 | 4.51±1.22 | 4.76±0.91 |
| 3 | 5.03±0.24 | 5.11±0.26 | 6.69±2.31 | 5.54±0.61 |
| 4 | 4.74±0.11 | 4.84±0.18 | 4.24±1.71 | 5.06±0.49 |
| 5 | 4.95±0.21 | 4.99±0.27 | 4.72±5.62 | 4.91±0.41 |
| 10 | N/A | 5.06 ± 0.07 | N/A | 5.09±0.28 |

**Table 2: Mean false-positive rates for different observation transformation functions $\mathcal{T}$.**

|  | $u_{a^*}$ | $R(f_{a^*})$ |
|---|---|---|
| EDA | 58.75 | 146.69 |
| PCA | 85.83 | 167.17 |
| IF | 86.74 | 217.94 |
| KNN | 87.38 | 188.22 |
| CBLOF | 131.00 | 231.61 |

**Table 3: The performance of different anomaly detectors for the exfiltration experiment. The models are sorted by attacker's best responce utility $u_{a^*}$.**

*6.1.2 Solution Quality.* Next, we compare the quality of computed classifiers. Since each randomly-generated instance may have a different game value, we use the concept of relative regret to compare the performance of the classifiers across different instances. We define the relative regret of two classifiers $c$ and $c_{base}$ as

$$\text{RelReg}(c, c_{base}) = 100 \frac{BR_a(c) - BR_a(c_{base})}{BR_a(c_{base})}, \quad (16)$$

where $BR_a(c) = \max_{f_a \in \mathcal{F}} u_a(c, f_a)$. The value $\text{RelReg}(c, c_{base})$ states how much the defender regrets choosing classifier $c$ compared to $c_{base}$. In case the value is negative, the quality of classifier $c$ is better compared to the baseline.

*Identity Transformation Function.* First, we compare EDA and LP in a simple scenario where the observation transformation function is an identity. In this case, we can closely approximate the optimal value using CFS. Figure 2(left) depicts relative regrets of LP and EDA compared to CFS. With the increasing dimension, the problem gets more complicated, thus the relative regret for each of the algorithm increases. However, while for the EDA the increase is not that dramatic and the mean value even for the feature space of dimension 20 is 6.46%, the relative regret for the LP-based method increases more dramatically, and even for dimension 5 exceeds 15%.

Next, we compare what is the false-positive rate of the algorithms on testing data. Table 2 shows the results, and both algorithms are able to find a classifier that keeps the desired false-positive rate close to 0.05.

*General Transformation Function.* For the more general case, we are unable to approximate the value of the game closely. Thus we compare only two algorithms – EDA and the LP-based method. Figure 2(right) depicts the relative regret showing that EDA produces significantly better classifiers compared to the LP-based method (the best response value of the attacker against EDA is significantly lower compared to the value against LP-based strategy). The main reason is that the discretization is much more coarse for higher dimensions. Moreover, since the LP has a hard constraint on the false-positive rate, the probability of detection for large cells created by the discretization has to be low and the attacker can exploit it more and achieve a better value. The second reason is that using KDE smoothing of input data can increase the density of benign data in some cells compared to original data. Comparison of the false-positive rates in Table 2 shows that due to coarse discretization in LP and transformation function, the EDA method is significantly better in maintaining the desired false-positive rate on testing data.

*6.1.3 EDA Convergence.* Figure 3 presents a typical learning of EDA during the training. There are two clear trends – first, the algorithm converges fairly quickly and already in iteration 250 the best response of the attacker decreases very close to the final value. Note that afterward, both the best response value of the attacker and the false-positive rate oscillate in opposite directions. This is expected since with increased false-positive rate the classifier is less exploitable and vice versa. Such an oscillation generates many points for the Pareto frontier from which the final classifier is selected (see Section 5.3).

## 6.2 Case Study: Comparison with Classical Anomaly Detectors over Data Exfiltration Problem

As the final experiment, we compare the performance of the EDA method in a real-world setting against standard anomaly detectors. We focus on the data exfiltration case via DNS protocol and use anonymized real-world 20,000 DNS queries. When the attacker is exfiltrating data via DNS queries, he splits data into parts and repeatedly encodes part of data as a part of the query (e.g., a third-level in a domain). We aggregate queries in a 20s time interval based on a common second-level domain. For simplicity, we consider only three features for this problem – the sum of lengths of all queries in a group within the time interval, the sum of entropy, and the number of special symbols. Reward function of the attacker should correlate with the amount of exfiltrated data and thus we approximate it as a multiplication of the first two feature values while the third feature does not affect the reward function.

We compare EDA with traditional anomaly detectors. We have chosen four well-known models: Principal Component Analysis (PCA), Isolation Forest (IF), K-nearest neighbors with largest (KNN) distances to the k-th neighbor as the outlier score and cluster-based local outlier factor (CBLOF) algorithm. We have used PyOD toolkit [30] with the implementations of these models.

Table 3 shows the results of all the classifiers sorted based on the performance against the best responding opponent (sample-based best response with 30, 000 samples). The results show that even in such a simple scenario, our EDA method outperforms all standard non-adversarial methods. The best non-adversarial anomaly detector, PCA, is much more exploitable compared to EDA and the expected best response utility of the attacker is almost 150% of the value against EDA. Moreover, the second column shows the reward values $R(f_{a^*})$ for the best-response action of the attacker – the higher the value, the better is the action for the attacker and this is the reward if this particular action is not investigated by the anomaly detector. Since EDA takes goals of the attacker into consideration, the classifier forces the attacker to choose best response actions with lower reward.

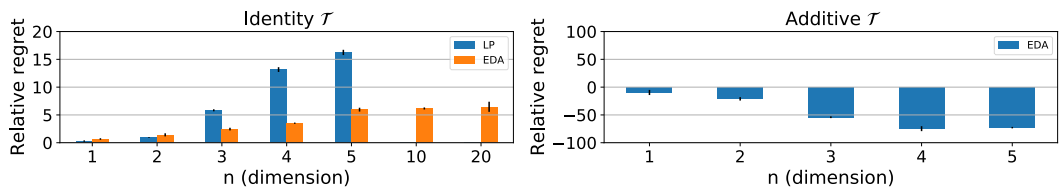

Figure 2: (left) Relative regret of LP and EDA compared to the optimal game-values computed with CFS identity transformation function $\mathcal{T}$ ; (right) Relative regret of EDA compared to LP for general transformation function $\mathcal{T}$.

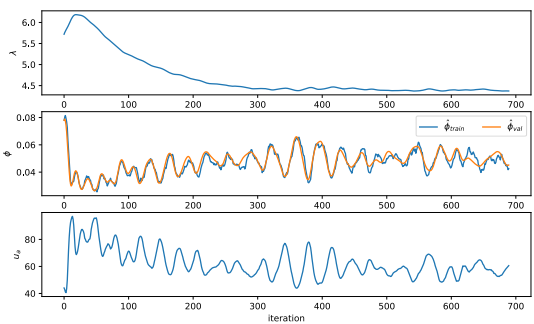

Figure 3: Visualisation of the convergence of variable $\lambda$, current false positive rate $\hat{\phi}^i$, and current best response value estimation $u_a$ for EDA during learning (the convergence is for the case study scenario).

## 7 CONCLUSION

We present a novel game-theoretic model for detecting anomalous events caused by the attacker. There are four key characteristics of our approach: the features considered by the anomaly detector can be continuous, there is a hard constraint for the expected false-positive rate of the detector, the defender cannot fully observe the action of the attacker, and the goal is to solve problems with multidimensional feature space. To address the computational challenges, we adapt a recent algorithm that trains a randomized classifier based on a neural network in order to minimize the exploitability. Experimental results show that our algorithm scales to larger dimensions compared to existing LP-based methods and also outperforms standard anomaly detectors on real-world DNS data exfiltration problem. Our work demonstrates that using game-theoretic algorithms, such as exploitability descent, can be used for adversarial anomaly detection but also adversarial machine learning in general and thus opens possibilities of adapting this approach to other domains.

## REFERENCES

[1] Ahmed, M.; Mahmood, A. N.; and Islam, M. R. 2016. A survey of anomaly detection techniques in financial domain. *Future Generation Computer Systems* 55:278–288.

[2] Athalye, A.; Carlini, N.; and Wagner, D. 2018. Obfuscated gradients give a false sense of security: Circumventing defenses to adversarial examples. In *International Conference on Machine Learning*, 274–283.

[3] Barreno, M.; Nelson, B.; Joseph, A. D.; and Tygar, J. D. 2010. The security of machine learning. *Machine Learning* 81(2):121–148.

[4] Bertsekas, D. 1999. *Nonlinear programming*. Athena Scientific optimization and computation series. Athena Scientific.

[5] Brückner, M., and Scheffer, T. 2011. Stackelberg games for adversarial prediction problems. In *Proceedings of the 17th ACM SIGKDD international conference on Knowledge discovery and data mining*, 547–555. ACM.

[6] Brückner, M.; Kanzow, C.; and Scheffer, T. 2012. Static prediction games for adversarial learning problems. *Journal of Machine Learning Research* 13(Sep):2617–2654.

[7] Chow, Y.; Ghavamzadeh, M.; Janson, L.; and Pavone, M. 2017. Risk-constrained reinforcement learning with percentile risk criteria. *The Journal of Machine Learning Research* 18(1):6070–6120.

[8] Dini, G.; Martinelli, F.; Saracino, A.; and Sgandurra, D. 2012. Madam: a multi-level anomaly detector for android malware. In *International Conference on Mathematical Methods, Models, and Architectures for Computer Network Security*, 240–253. Springer.

[9] Dritsoula, L.; Loiseau, P.; and Musacchio, J. 2017. A game-theoretic analysis of adversarial classification. *IEEE Transactions on Information Forensics and Security* 12(12):3094–3109.

[10] Durkota, K.; Lisý, V.; Kiekintveld, C.; Horák, K.; Bošanský, B.; and Pevný, T. 2017. Optimal strategies for detecting data exfiltration by internal and external attackers. In *International Conference on Decision and Game Theory for Security*, 171–192. Springer.

[11] Eykholt, K.; Evtimov, I.; Fernandes, E.; Li, B.; Rahmati, A.; Xiao, C.; Prakash, A.; Kohno, T.; and Song, D. 2017. Robust physical-world attacks on deep learning models. *arXiv preprint arXiv:1707.08945*.

[12] Garcia-Teodoro, P.; Diaz-Verdejo, J.; Maciá-Fernández, G.; and Vázquez, E. 2009. Anomaly-based network intrusion detection: Techniques, systems and challenges. *computers & security* 28(1-2):18–28.

[13] Goodfellow, I. J.; Shlens, J.; and Szegedy, C. 2014. Explaining and harnessing adversarial examples. *arXiv preprint arXiv:1412.6572*.

[14] Huang, L.; Joseph, A. D.; Nelson, B.; Rubinstein, B. I.; and Tygar, J. 2011. Adversarial machine learning. In *Proceedings of the 4th ACM workshop on Security and artificial intelligence*, 43–58. ACM.

[15] Krizhevsky, A.; Sutskever, I.; and Hinton, G. E. 2012. Imagenet classification with deep convolutional neural networks. In Pereira, F.; Burges, C. J. C.; Bottou, L.; and Weinberger, K. Q., eds., *Advances in Neural Information Processing Systems 25*. Curran Associates, Inc. 1097–1105.

[16] Lisý, V.; Kessl, R.; and Pevný, T. 2014. Randomized operating point selection in adversarial classification. In *Joint European Conference on Machine Learning and Knowledge Discovery in Databases*, 240–255. Springer.

[17] Lockhart, E.; Lanctot, M.; Pérolat, J.; Lespiau, J.-B.; Morrill, D.; Timbers, F.; and Tuyls, K. 2019. Computing approximate equilibria in sequential adversarial games by exploitability descent. *arXiv preprint arXiv:1903.05614*.

[18] Madry, A.; Makelov, A.; Schmidt, L.; Tsipras, D.; and Vladu, A. 2017. Towards deep learning models resistant to adversarial attacks. *arXiv preprint arXiv:1706.06083*.

[19] Moosavi-Dezfooli, S.-M.; Fawzi, A.; and Frossard, P. 2016. Deepfool: a simple and accurate method to fool deep neural networks. In *Proceedings of the IEEE*

*conference on computer vision and pattern recognition*, 2574–2582.

[20] Nair, V., and Hinton, G. E. 2010. Rectified linear units improve restricted boltzmann machines. In *Proceedings of the 27th International Conference on International Conference on Machine Learning*, ICML'10, 807–814. USA: Omnipress.

[21] Nelson, B.; Barreno, M.; Chi, F. J.; Joseph, A. D.; Rubinstein, B. I.; Saini, U.; Sutton, C.; Tygar, J.; and Xia, K. 2009. Misleading learners: Co-opting your spam filter. In *Machine learning in cyber trust*. Springer. 17–51.

[22] Perolat, J.; Malinowski, M.; Piot, B.; and Pietquin, O. 2018. Playing the game of universal adversarial perturbations. *arXiv preprint arXiv:1809.07802*.

[23] Shoham, Y., and Leyton-Brown, K. 2008. *Multiagent systems: Algorithmic, game-theoretic, and logical foundations*. Cambridge University Press.

[24] Sommer, R., and Paxson, V. 2010. Outside the closed world: On using machine learning for network intrusion detection. In *2010 IEEE symposium on security and privacy*, 305–316. IEEE.

[25] Szegedy, C.; Zaremba, W.; Sutskever, I.; Bruna, J.; Erhan, D.; Goodfellow, I.; and Fergus, R. 2013. Intriguing properties of neural networks. *arXiv preprint arXiv:1312.6199*.

[26] Thomas, K.; McCoy, D.; Grier, C.; Kolcz, A.; and Paxson, V. 2013. Trafficking fraudulent accounts: The role of the underground market in twitter spam and abuse. In *Presented as part of the 22nd USENIX Security Symposium (USENIX Security 13)*, 195–210.

[27] Vorobeychik, Y., and Kantarcioglu, M. 2018. Adversarial machine learning. *Synthesis Lectures on Artificial Intelligence and Machine Learning* 12(3):1–169.

[28] Wang, G.; Wang, T.; Zheng, H.; and Zhao, B. Y. 2014. Man vs. machine: Practical adversarial detection of malicious crowdsourcing workers. In *23rd USENIX Security Symposium (USENIX Security 14)*, 239–254.

[29] Wong, E., and Kolter, Z. 2018. Provable defenses against adversarial examples via the convex outer adversarial polytope. In *International Conference on Machine Learning*, 5283–5292.

[30] Zhao, Y.; Nasrullah, Z.; and Li, Z. 2019. Pyod: A python toolbox for scalable outlier detection. *Journal of Machine Learning Research* 20(96):1–7.